\documentclass[letterpaper, 10 pt, conference]{ieeeconf}  

\IEEEoverridecommandlockouts                              

\usepackage{epsfig} 

\title{\LARGE \bf
MLR (Memory, Learning and Recognition): A General Cognitive
Model - applied to Intelligent Robots and Systems Control
}

\author{Aras R. Dargazany 
\thanks{Aras Dargazany, Department of Electrical, Computer, and Biomedical Eng., Univ. Rhode Island {\tt\small arasdar@uri.edu}}
}

\begin{document}

\maketitle
\thispagestyle{empty}
\pagestyle{empty}

\begin{abstract}

This paper introduces a new perspective of intelligent robots and systems control. 
The
presented and proposed cognitive model: 
Memory, Learning and Recognition (MLR), is an effort to bridge
the gap between Robotics, AI, Cognitive Science, and
Neuroscience. 
The currently existing gap prevents us from integrating the current advancement and achievements of
these four research fields which are actively trying to define
intelligence in either application-based way or in generic
way.

This cognitive model defines intelligence more specifically,
parametrically and detailed.
The proposed MLR model helps
us create a general control model for robots and
systems independent of their application domains and platforms
since it is mainly based on the dataset provided
for robots and systems controls. 

This paper is mainly proposing
and introducing this concept and trying to prove this concept
in a small scale, firstly through experimentation. 
The
proposed concept is also applicable to other different platforms
in real-time as well as in simulation.

\end{abstract}

\section{Introduction}
\label{Introduction}

Intelligent control of robots and systems have been at the center
of attention for decades since Robotics is a very
useful and applicable field of research in human life. 
After DARPA 2005 \cite{Thrun_stanley, Thrun_junior, Thrun_toward_robots}, there have been significant
investments into Probabilistic Robotics \cite{thrun_book} (i.e. the usage of learning for robot control) and more specifically, in Autonomous Navigation.

Since then, we have also seen much industrial inclination toward applying this technology
to their own products to create more intelligent machines
such as Google self-driving cars. 
The main drawback, however,
has always been a lack of common intelligent architecture
or a lack of a concrete and applicable definition of intelligence
which should somehow explain how it is possible to create an
intelligent system. 

Although there has been an enormous
amount of financial investment with many attempts to define intelligence from different fields of research,
none of them can actually explain intelligence (human
intelligence specifically) in a way that it can be applicable
to other fields of research, thus connecting all, i.e. there is a lack of an operational definition. 

The current advancements in machine learning
known as Deep Learning \cite{LeCun_15_nature} has brought much hope
and excitement about AI general applications and perhaps this is the right way to approach intelligence
and its definition \cite{true_intelligence}. 
An interesting review paper presents the
complete historical trend of the deep learning approach in Neural
Networks \cite{DL_review}. 

Another
recent excitement in AI was brought by the Deep Reinforcement
Learning method proposed by \cite{DeepMind_15_nature} which is presenting a Deep
Learning Approach in Atari games which is also a very
simple and primitive simulated environment and robots (so-
called agent in the paper) for testing the control efficiency
of the deep learning approach. 

The latter proposed approach
claims to outperform human-level control of the simulated
agent in the simulated environment of the Atari games; 
this 
is a very promising step forward toward understanding the
true nature of intelligence and intelligent control of robots.
The latter approach, \textbf{DQN}, has triggered some
researchers to look into the new theory of \textit{Learn-See-Act}
instead of the previous \textit{Sense-Plan-Act} \cite{Learn_See_Act} and \cite{learning_behaviours}.

\section{MLR (Memory, Learning and Recognition)}
\label{MLR} 
\begin{figure}[thpb]
    \centering
    \includegraphics[scale=0.4]{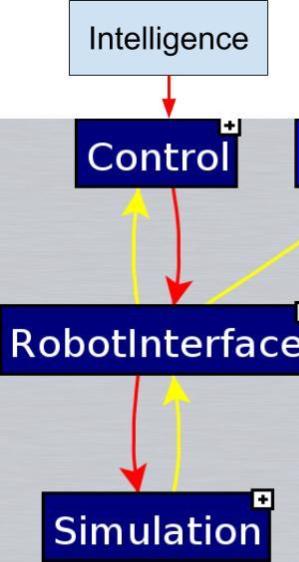}
    \includegraphics[scale=0.4]{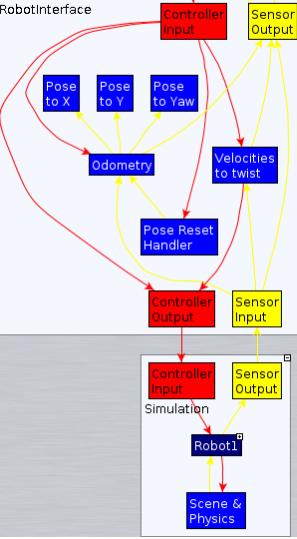}
    \caption{(left)The robot control architecture with four layers is presented, (right) Inside the first two layers are also shown.}
    \label{figure:architecture}
\end{figure}

The proposed generic cognitive model is basically suggesting three main components for intelligent control: Memory,
Learning and Recognition.
These three components can also be categorized into three: \textbf{Cognition (done by Recognition), knowledge (stored in Memory) and Intelligence (produced by learning)} 
(please be advised that this categorization is proposed by this paper). 
Intelligence includes a combination of Memory and Leaning. 
For better understanding of the proposed model, the complete intelligent robot control architecture is illustrated
in figure ~\ref{figure:architecture}.

The left figure in \ref{figure:architecture} is showing that a complete intelligent robot and systems control architecture has four main layers; 
The first two layers are shown in more details in the right figure in ~\ref{figure:architecture}.

As you can see the first two layers are preparing the robot to be able to deliver the sensor data to control module and also be able to receive the controlling commands from the control module in the third layer.
It is important to know that our proposed cognitive model (MLR) is focusing only on the last and highest two layers, which defines intelligent control.

\begin{figure}
    \centering
    \includegraphics[scale=0.3]{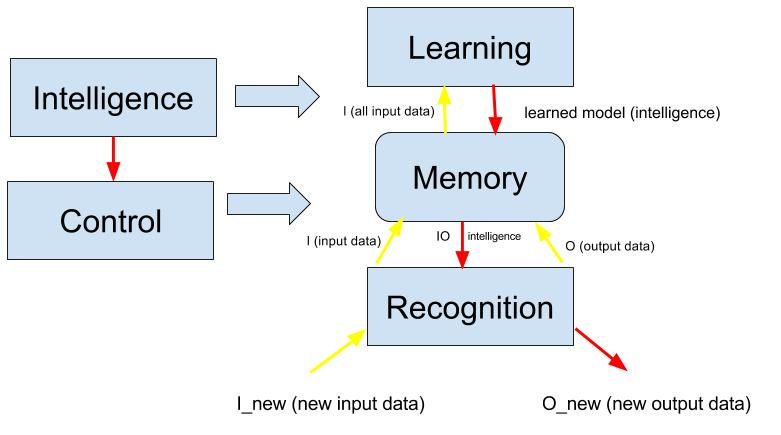}
    \caption{The main proposed conceptual cognitive model of intelligent control introduced as \textbf{MLR}}.
    \label{figure:MLR_cognitive_model}
\end{figure}

The first two layers are provided and illustrated along with the last and highest two layers to give you an overview of how our complete robot control architecture looks like and also
how many layers are needed in either real or simulated robot in order to create a complete intelligent robot control.
Figure \ref{figure:MLR_cognitive_model} gives an overview of what is proposed in this paper
as the cognitive model.

\begin{figure}
  \centering
 \includegraphics[scale=0.25]{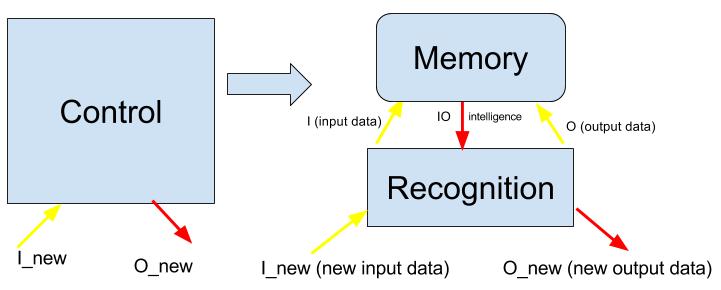}
 \caption{The separate control module.}
 \label{figure:control}
\end{figure}

\begin{figure}
  \centering
 \includegraphics[scale=0.25]{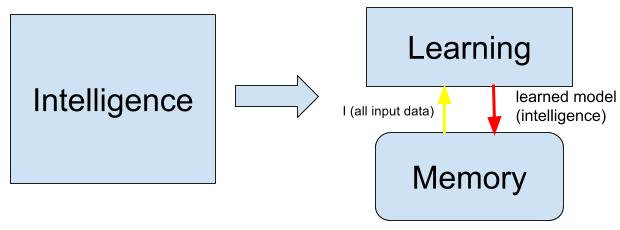}
 \caption{The separate intelligence module.}
 \label{figure:intelligence}
\end{figure}

Based on this model, the control module equals to memory
and recognition but the intelligence module is composed of
memory and learning. The interesting thing about this model
is that these two processes are done completely independently
and can be performed on different computers and processors. 
In figures \ref{figure:control} and \ref{figure:intelligence}, you can see the model in more
detail and as two completely independent processes.

After Looking at the overall MLR architecture (the proposed cognitive model for intelligent control), now we will
get into more detail on each one and how
they work together.

\subsection{Memory}
\label{subsec:memory}

Memory is creating the data space and storage needed
for writing data and reading data. Based on the memory
limit and therefore our database limit, we can conclude
that the intelligence created using the learning module and
by learning the stored data in memory can not be beyond
the limits of our input and output data stored in our
memory. Basically, memory is giving us the database
and based on the database, we can begin to create our own
data space using learning. It is also very important to know
that memorizing and remembering things play an important
role in human intelligence and so the same may apply
for robots and intelligent systems as well. In the proposed
concept, we will start with writing Input Output $IO$ data to
our memory (or the Hard disk of the computer).
$$
IO = \left \{ IO_{1},............., IO_{t} \right \} \eqno{(1)}
$$

These input output data are sampled throughout time and
are generally \textbf{Sensor Output} $I$ and \textbf{Controller Input} $O$
as shown in Ò(1). Parameter $t$ is the time index in which
the data samples were acquired and recorded into memory.

The recorded $IO$ database in the memory is completely
based on the manual controlling of the robot. Therefore,
it is highly suggested to create Input-based Output, meaning
the Sensor-based control data which is going to be used later 
by the robot itself for intelligent control.

\subsection{Learning}
\label{subsec:learning}
Having written and prepared the $IO$ database, we will
use all the input data I for learning. Learning, in
our work, is analyzing the input data space, decomposing data
eigenvalues for creating input data eigenspace finally. 
But the main idea in learning is usually creating the Data Space
in the first place. Data Space is also known as Data Feature Space
or Feature Space. In order to create the feature space, We
need to read all the input data from the memory as mentioned in Ò(2) and
create a matrix of all input data vectors all at once.
$$
I = \left \{ I_{1}, ....,I_{t} \right \} \eqno{(2)}
$$

Once we have the matrix $I$, we mathematically have created
the input data feature space. This matrix is a \textbf{Column-Major matrix}
meaning that the number of columns are indicating the number of input data samples and the
number of rows are indicating the number of input data dimensions.

Using PCA helps us create the data feature eigen space 
which is basically composed of the principle components. 
Principle components are also known as
eigen values and eigen vectors of $I$ \textbf{(Input Data Matrix)}.
In order to do PCA, First of all we have to calculate $\mu$ (the
mean) of $I$ as Òequation (3):
$$
\mu = \frac{\sum_{1}^{t} I_{i}}{t}  \eqno{(3)}
$$

Given $\mu$, we will start translating all the input vectors to
the origin as Òequation (4):
$$
I - \mu= \left \{ I_{1} - \mu, ....,I_{t} - \mu \right \} \eqno{(4)}
$$
$$
I - \mu = \varphi,\ I_{1} - \mu = \varphi_{1},........,\ I_{t} - \mu = \varphi_{t}
$$
$$
\varphi = \left \{ \varphi_{1}, ......., \varphi _{t} \right \}
$$

Once we have all $\varphi$, now we can start the principle component analysis (PCA) or eigenvalue decomposition as shown in ~\ref{figure:intelligence} and mentioned in following Òequation (5):
$$
A\cdot \nu = \lambda \cdot \nu \Rightarrow A = \nu \cdot \lambda \cdot \nu ^{t} \eqno{(5)}
$$

Using eigenvalue decomposition, we can measure eigenvalues $\lambda$ and eigenvectors $\nu$. Also we should not forget that,
according to Singular Value Decomposition (SVD) Òequation (5) and (6),
$$
\varphi = \nu \cdot \Lambda \cdot \upsilon ^{t} \Rightarrow \lambda = \Lambda ^{2} \eqno{(6)}
$$
$$
\varphi \cdotp \varphi^{t} = (\nu \cdot \Lambda \cdot \upsilon ^{t}) \cdotp (\nu \cdot \Lambda \cdot \upsilon ^{t})^{t} = \nu \cdot \Lambda^{2} \cdot \nu^{t}
$$

\begin{figure}
    \centering
    \includegraphics[scale=0.39]{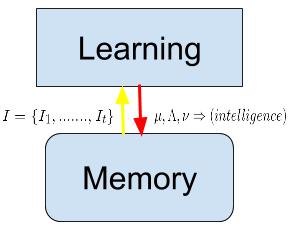}
    \caption{The input and output of the learning module (data flow).}
    \label{figure:learning}
\end{figure}

Therefore using the equations above Òequation (5) and (6), singular values $\Lambda$ can be measured using eigenvalues
$\lambda$ which will help us in scaling that will be used later for recognition purposes ~\ref{subsec:recognition}. 
Having calculated $\mu$, $\Lambda$, $\nu$, we write them back
into the memory. Basically the learning module reads the
recorded input data from memory $I$ and generates the learned
eigenspace information such as: $\mu$, $\Lambda$, $\nu$ which can be used
to reduce the data dimensionality considerably to the handful
of principle components. 
In figure ~\ref{figure:learning}, the input and
the output of the learning module are specifically shown in
reading $I$ from memory and writing $\mu$, $\Lambda$, $\nu$ back into the
memory.

\subsection{Recognition}
\label{subsec:recognition}

\begin{figure}
 \centering
 \includegraphics[scale=0.3]{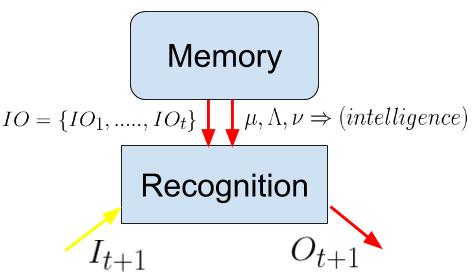}
 \caption{The input and output of recognition module (data flow).}
 \label{figure:recognition}
\end{figure}

Having learned the input data as discussed in previous
section \ref{subsec:learning}, this module read the $IO$ recorded data both
from sensors and controllers along with $\mu, \Lambda, \nu$ (learned
principle components) as shown
in figure ~\ref{figure:recognition}.

Once we read all the recorded data from the memory with their time index $IO_{t}$, we can start comparing the new input data $I_{t+1}$ with our learned input data in our database $I$.
This comparison $I_{t+1}$ vs $I={I_{1},......, I_{t}}$ should be done intelligently meaning using the principle components ($\mu, \Lambda, \nu$) (or intelligence parameters).
Basically in recognition, we are comparing $I_{t+1}$ with every single one of the input data in $I$ based on four different metrics in order to find the \textbf{most similar}
and the \textbf{least different} existing recorded input data in our learned database.

Once we read all the recorded data from the memory with
their time index $IO_{t}$, we can start comparing the new input
data $I_{t+1}$ with our learned input data in our database $I$. This
comparison $I_{t+1}$ vs $I={I_{1},......, I_{t}}$ should be done intelligently meaning using the principle components ($\mu, \Lambda, \nu$)
(or intelligence parameters). Basically in recognition, we are
comparing $I_{t+1}$ with every single one of the input data in
$I$ based on four different metrics in order to find the most
similar and the least different existing recorded input data
in our learned database. In order to compare them, at first we need to project them onto the learned eigen space using $\mu, \Lambda, \nu$ as follows:
$$
\varphi_{t+1} = I_{t+1} - \mu,\ \varphi = \left \{ \varphi_{1},......., \varphi_{t} \right \} \eqno{(7)}
$$

It is also important to know that the resulting number of principle components in maximum can be equal to the number of data samples $t$ but since we want to reduce the dimensionality of our data in a way 
that we can still accurately enough be able to reconstruct them, therefore $\nu = \left \{ \nu_{1},........, \nu _{n} \right \}$, meaning that
we only keep $n$ number of principle components ($\frac{n}{t}$ percentage of kept eigenvectors).

In Òequation (7), $n$ is indicating the number of principle components which is also \textit{considerably reduced} compared to the initial number of input data dimensions.
$$
\omega_{\varphi_{1}} = \nu \cdot \varphi_{1}, ......................, \omega_{\varphi_{t}} = \nu \cdot \varphi_{t} \eqno{(8)}
$$
$$
\omega_{\varphi_{t+1}} = \nu \cdot \varphi_{t+1} \eqno{(9)}
$$ 

In Ò(8) and Ò(9), both new input data $I_{t+1}$ and the recorded input data in memory $I$ are projected to the learned eigenspace and can be reconstructed as shown in Òequation (10):
$$
\varphi_{t+1} = \omega_{\varphi_{t+1}} \cdot \nu \eqno{(10)}
$$

When you can completely reconstruct all the recorded input data $I$ and the new one $I_{t+1}$,
this means that we can easily compare them using their new projected values $\omega_{\varphi}$ vs $\omega_{\varphi_{t+1}}$ in the eigen space based on their 
principle components $\Lambda\nu = \left \{ \Lambda_{1}\nu_{1},................., \Lambda_{n}\nu_{n} \right \}$ based on \textbf{\textit{four metrics}} as follows:

\subsubsection{Minimum Squared Difference}
\label{subsubsec:MSD}
This metric is based on minimum distance between two vectors in the eigenspace as shown in Òequation (11). 
This metric is also known as \textbf{Min Square Error} and the min is 0 which means that the two vectors have no difference in this space and the max can be any value.
$$
\left | \omega_{\varphi_{t+1}} - \omega_{\varphi_{1}}  \right |,..................................., \left | \omega_{\varphi_{t+1}} - \omega_{\varphi_{t}}  \right |  \eqno{(11)}
$$

\subsubsection{Scaled Minimum Squared Difference} 
\label{subsubsec:SMSD}
Minimum scaled difference is also based on minimum distance \textbf{but} it also applies the importance of every single principle components and 
since $\Lambda$ is basically $\omega_{max}$ on each component $\nu$, that is why using $\Lambda$ for scaling might really make the comparison more fair and more accurate 
(at least \textit{theoretically}) between two vectors in the space as shown in Òequation (12).
$$
\left \{ \left | \frac{\omega_{\varphi_{t+1}}}{\Lambda} - \frac{\omega_{\varphi_{1}}}{\Lambda} \right |,........, \left | \frac{\omega_{\varphi_{t+1}}}{\Lambda} - \frac{\omega_{\varphi_{t}}}{\Lambda} \right |  \right \}  
\eqno{(12)}
$$

\subsubsection{Maximum Normalized Cross Similarity}
\label{subsubsec:MNCS}
This metric is mainly measuring the angle between two vectors in the space and since the smaller the angle is, the more similar these two vectors are. 
This metric is also known as \textbf{Normalized Cross Correlation} which is basically the dot product or inner product of two vectors in the space 
as shown in Òequation (13) as follows:
$$
\left \{ \frac{\omega_{\varphi_{t+1}}}{\left | \omega_{\varphi_{t+1}} \right |} \cdot \frac{\omega_{\varphi_{1}}}{\left | \omega_{\varphi_{1}} \right |},..................., \frac{\omega_{\varphi_{t+1}}}{\left | \omega_{\varphi_{t+1}} \right |} \cdot \frac{\omega_{\varphi_{t}}}{\left | \omega_{\varphi_{t}} \right |} \right \}
 \eqno{(13)}
$$

The result of this metric is always $0 \leq cos\angle \leq 1$ which makes it work more like a probability measurement.

\subsubsection{Scaled Maximum Cross Similarity}
\label{subsubsec:SMCS}
This metric is also mainly based on previous metric \textbf{with two main differences}:
\begin{itemize}
 \item it is not normalized
 \item it is scaled using $Lamda$ with completely the same scaled minimum squared difference. 
\end{itemize}
As you can see it in Òequation (14) as follows:
$$
\left \{ \frac{\omega_{\varphi_{t+1}}}{\Lambda}\cdot \frac{\omega_{\varphi_{1}}}{\Lambda}, ....................., \frac{\omega_{\varphi_{t+1}}}{\Lambda}\cdot \frac{\omega_{\varphi_{t}}}{\Lambda} \right \}
 \eqno{(14)}
$$

Using all four metrics, we can almost precisely find the most similar and least different input data among $I = \left \{ I_{1},......................, I_{t} \right \}$ to 
the new input data $I_{t+1}$ and once it is found, we can use its index $t$ and we can use its corresponding recorded output $IO = \left \{ IO_{1},......................, IO_{t} \right \}$
as the new output $O_{t+1}$ for the new input data $I_{t+1}$.

\section{Experimental results}
\label{sec:experiment}
In order to speed up the implementation of the proposed MLR model, we decided to use the \textbf{simulation}.
The simulated environment and simulated robot is the default, well-maintained and well-documented project in Finroc \cite{finroc}.
For the experimental setup, we are using Linux kernel in Ubuntu 14.04 (64 bit) and also as explained above, Finroc is our framework which is using its own simulation environment 
known as SimVis3D. The current available open-source project in Finroc is called Forklift and it has been used as the simulated experimental platform for 
testing the performance of our MLR model as follows:
We chose our experimental setup in a way that it can show the exact role of each module in it and how they are working together.

\subsection{Recording the dataset into memory}
\label{subsec:experimental_memory}
The memory is very important module in the proposed MLR model since we have to start managing our memory in our experiments at first by writing sensor data into memory (will be later used as $I$).
These sensor data are specifically camera images (the highest possible image resolution in simulation $900 \times 700$ size and in $RGB$ scale), distance data and localization data.
In this experiment, we decided to use only the camera images to show the power of the MLR model in dealing with high-dimensional input data and also for more clear understanding of the performance of the model
in generating the controlling commands to the robot.

Also, we should write the controlling data $O$ at the same time with sensor data (controller data are generated initially by the data recorder and the person who is manually controlling the robot to record the data). 
These controlling commands are $O$ which will be recorded along with $I$ at the same time and gives us $IO$ dataset in the memory.
The writing and reading data from memory has been illustrated before in figures ~\ref{subsec:learning}, ~\ref{subsec:memory} and ~\ref{subsec:recognition}.

\begin{figure}
 \centering
 \includegraphics[scale=0.14]{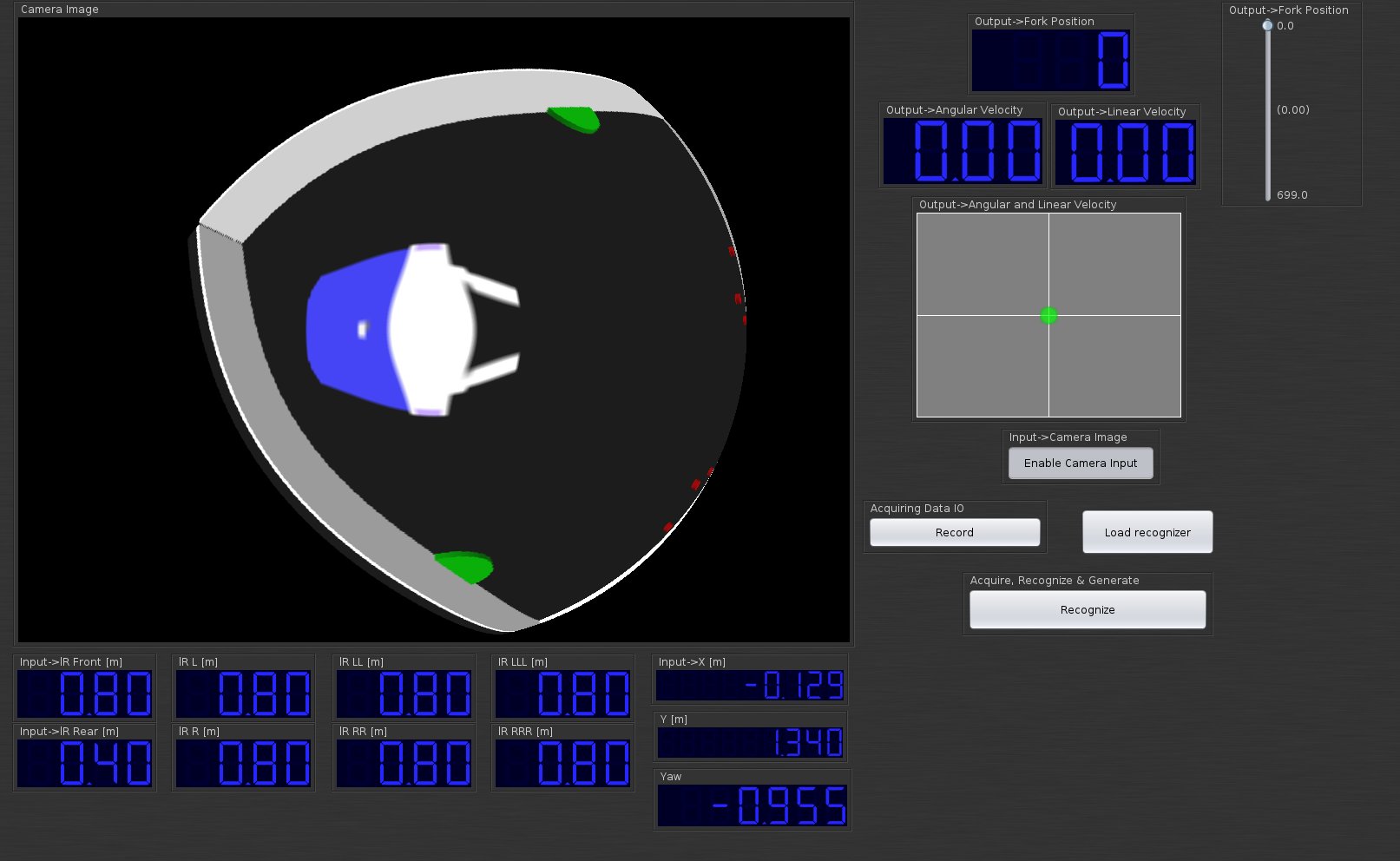}
 \caption{A snapshot taken from GUI used to manually control the robot and record the $IO$ dataset.}
 \label{figure:fingui}
\end{figure}

\begin{figure}
 \centering
 \includegraphics[scale=0.45]{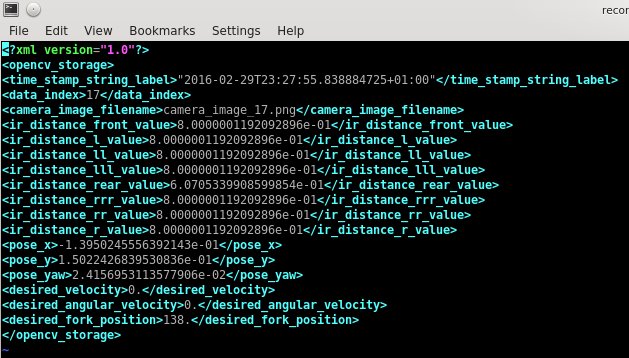}
 \caption{The XML format in which the recorded data are converted to along with their labels and directory for better and accurate recognition.}
 \label{figure:XML_data}
\end{figure}

Using the Finroc GUI (Known as FinGUI), we can manually control the robot using joystick to explore the default simulated environment as shown in figure \ref{figure:fingui}.
At every specific date and specific length of time (or duration of time) in that date, we record only one dataset $IO = \left \{ IO_{1},.................,IO_{t} \right \}$. 
The date of recording will be used as the directory in which all the data recorded will be written to and during recording, one index will be assigned to all of the data.
The date of recording and the index of recorded data both are stored in an XML file shown in figure \ref{figure:XML_data}.

As you can see in this figure, the first row is \textbf{time stamp label string} which is the exact date of data recording accuracy.
The second row is \textbf{data index} which is assigned to the recorded data during the recording to keep corresponding input and output data together.
Data index also depends on the duration of recording the data as well. 
In this experiment, we had two different recordings which means two different time stamps or two complete different datasets.
Each one of these recordings took about 10 minutes and we were recording at the speed on 1 data per second.

This means we have generated $1 \times 10 \times 60 = 600$ indexes of data in each recording or $IO = {IO_{1},..........,IO_{600}}$.
Therefore, in one recording we have a total of 600 camera images, with resolution of $900 \times 700$ size and in $RGB$ scale recorded in our memory.
These camera images will be used for learning and also for recognition.
In XML file \ref{figure:XML_data}, the third row is also the name of the camera image and this image is stored in the same folder as the XML file is stored.
The name of the folder (or the parent folder) is the same as the time stamp label which is the date of recording.

Also in XML file \ref{figure:XML_data}, after the camera image, there are distance data value which are more specifically 8 infra-red distance data values and 
localization data is also place after distance values composed of pose $X, Y and\ Yaw$.
After localization data, there are controlling data value generated at the same time as these sensor data values.
Controller value are desired linear velocity and angular velocity of the robot and also the position robot's fork.

As you can see OpenCV \cite{opencv} cv::FileStorage framework has been used for storing data into the memory.
Also boost filesystem library \cite{boost} is used to search for files and store their paths so that it is possible to read the files from directory based on the XML file \ref{figure:XML_data}.

It is also important to notice that sensor data and controller data depends on the person who is manually controlling the robot and recording the data.
Finally these recorded data create the robot intelligence which means that it is highly dependent on the data recorder intelligence.

In figure \ref{figure:fingui}, the GUI for manually controlling the simulated robot and recording the data are shown along with the controller values on top of joystick and 
the distance value and localization values and camera image at the left side of the joystick. 
At right side of the joystick you can also see the fork position slider to manipulate the objects and obstacles in the environment.

\subsection{Learning the recorded dataset}
\label{subsec:experimental_learning}
Having recorded the data in the memory with labels including date of recording and the index of the recorded, now the learning process has begun. 
The tricky part here is to choose a threshold for the number of eigenvalues and the maximum number of eigenvalues to keep without
hurting the accuracy of the work and also it is going to be very slow during the recognition process.

In this experiment, we are only using the camera images with highest possible resolution in Finroc simulation framework SimVis3D in order to prove the concept in handling a very large dimensional input data.
Having read the file path of each camera images, we store them first into OpenCV Matrix and then we push them one by one to the vector. 
Basically each Matrix is the image data structure used in OpenCV \cite{opencv} and 
the vector is the standard library vector class which is very efficient as data container class.

Once all the camera images are read and collected, we convert them all intro gray scale since they are all in RGB format.
Therefore, we have to change them all from RGB (3-channel) to grayscale (1-channel) and 
also make sure that they are normalized meaning the pixel values are between 0-255.
I call this process scaling the data and it is mainly composed of converting RGB to Gray scale and normalizing the data to 0-255.

Having scaled the data, now we should vectorize all of our images or in simpler way we should change all the images into one column image.
In result, we change vector to only one Mat in whose each column there is only one image. Images are stored as Column Data.
This Matrix (Mat) is called $I = {I_{1},........,I_{t}}$. In this case $t = 600$ as explained in the process of recording dataset into memory \ref{subsec:experimental_memory}.
Having created $I$ all the input data matrix (images in this case), now data space can be created and learned using PCA engine in OpenCV \cite{opencv}.
There are plenty of implementations of PCA, SVD and Eigen Value Decomposition (EVD) which might be implemented in different ways.

\begin{figure}
 \centering
 \includegraphics[scale=0.13]{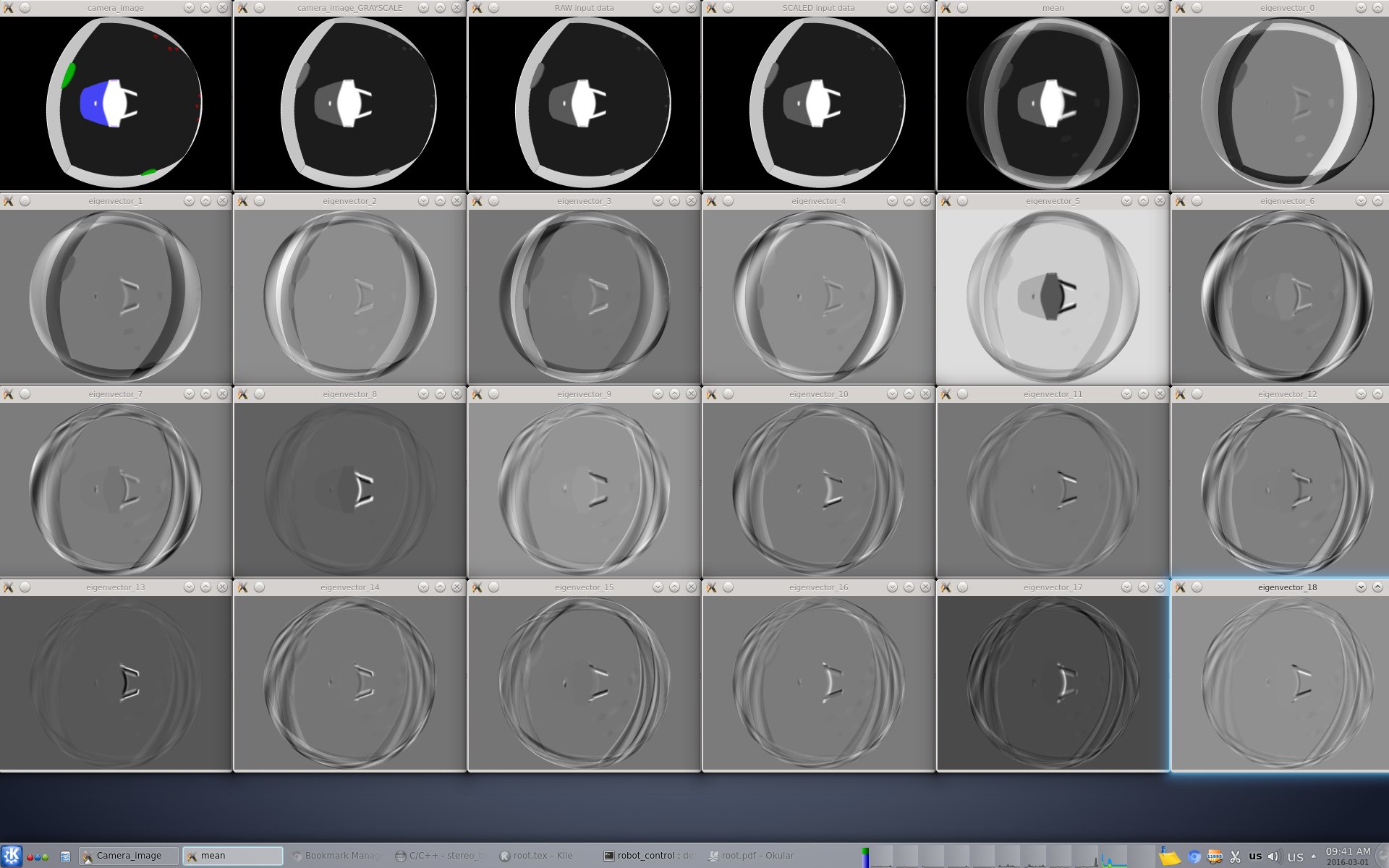}
 \caption{The images learned and the resulting mean $\mu$ and eigenvectors $\nu$ are all visualized.}
 \label{figure:learned_images}
\end{figure}

\begin{figure}
 \centering
 \includegraphics[scale=0.15]{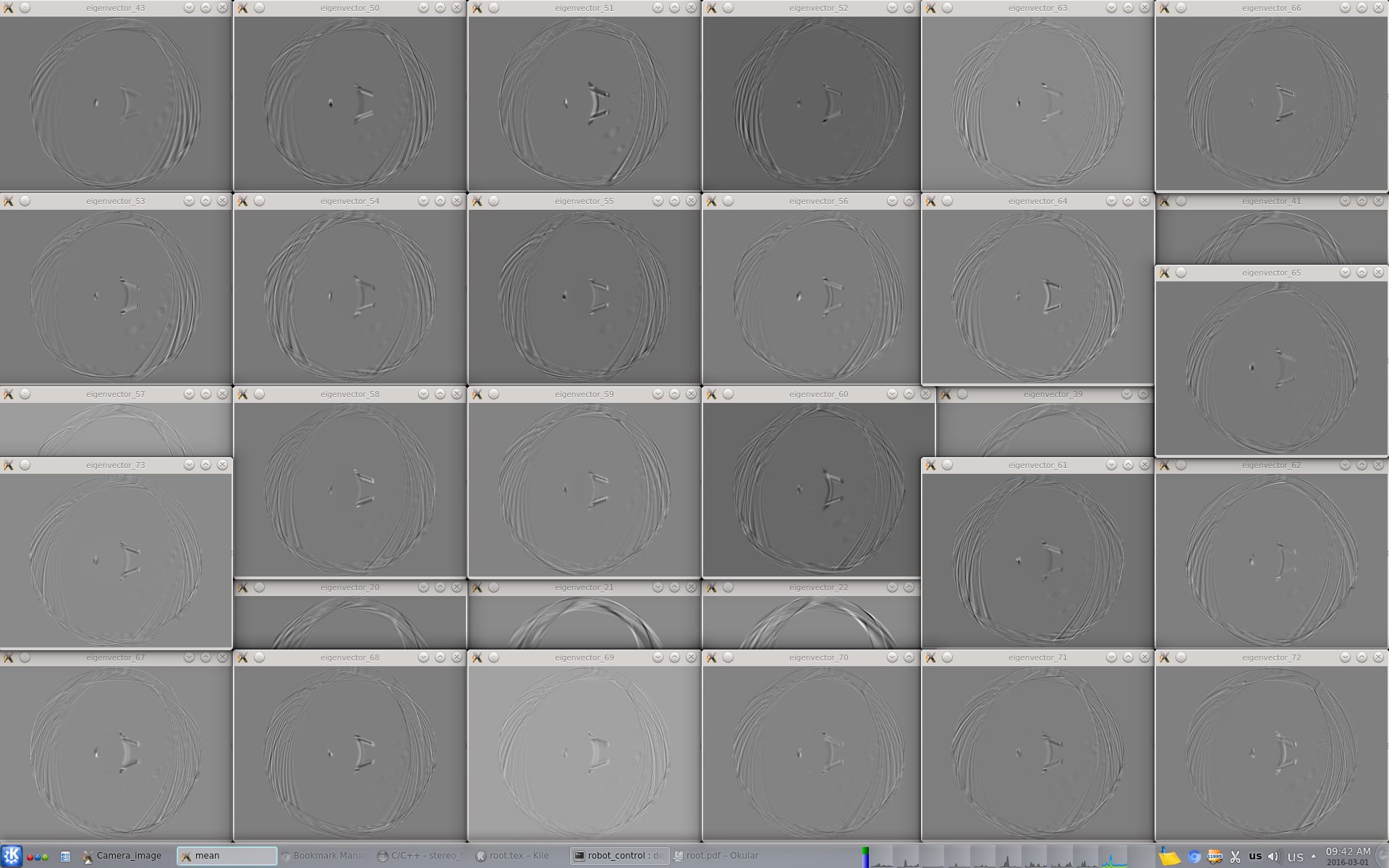}
 \caption{The rest of eigenvectors corresponding to smaller eigenvalues are shown to give you an overview of how these eigenvectors tend to vary.}
 \label{figure:more_eigenvectors}
\end{figure}

\begin{figure}
 \centering
 \includegraphics[scale=0.15]{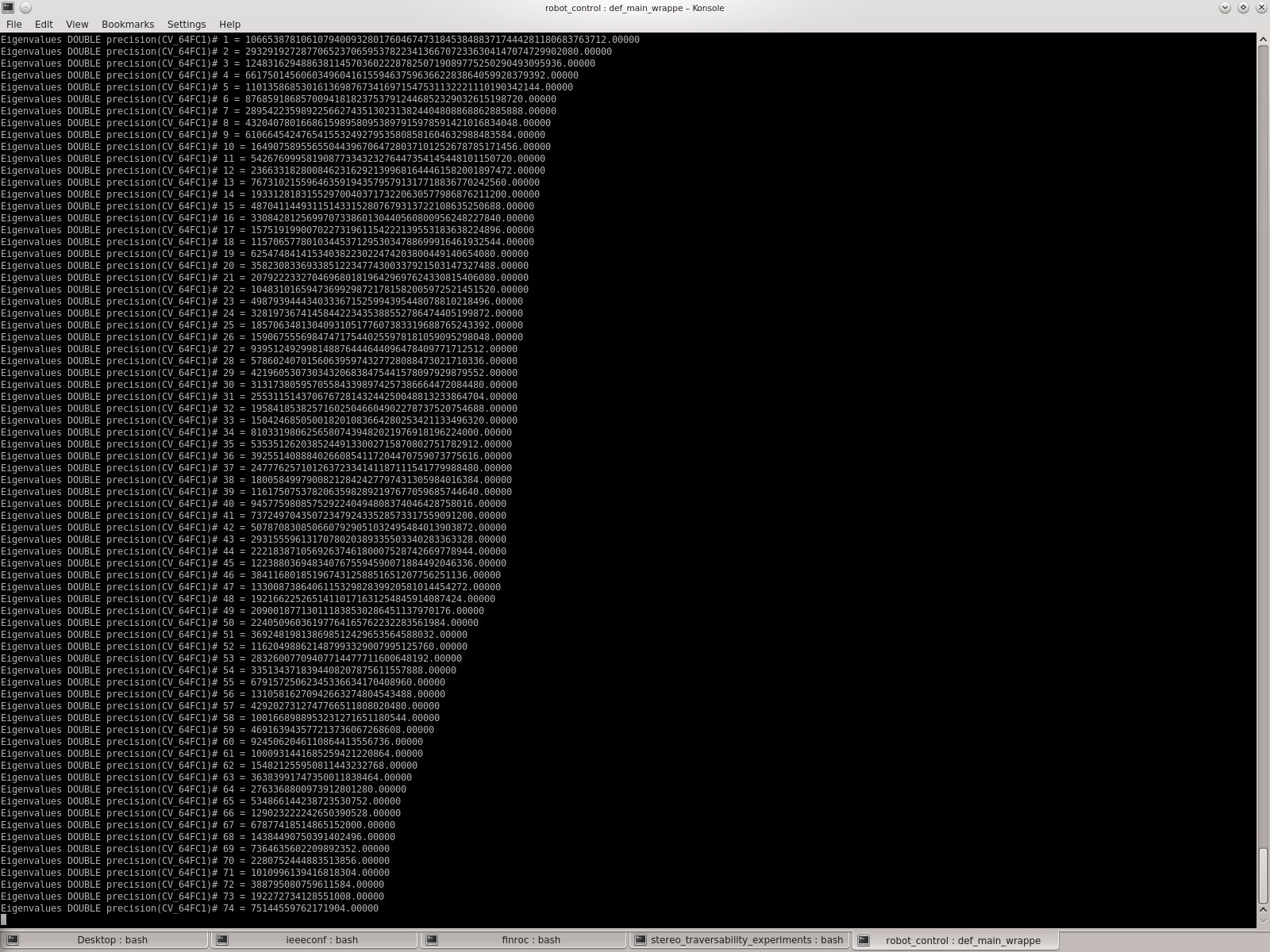}
 \caption{The eigenvalues are shown in order of their values and the trend of change is also clear in the length their value from top to bottom (they are completely descending).}
 \label{figure:eigenvalues}
\end{figure}

This would be an interesting research idea to also compare their results all together and see if their results are the same or not since
some of them like OpenCV PCA engine is supporting the float precision value and some other implementations such as Matlab and Python are generating the results with Double precision.
Using the PCA engine, we can calculate the mean $\mu, \Lambda and \nu$ which are all shown in figure \ref{figure:learned_images} 
in order of the eigenvalues which are also shown in figure \ref{figure:eigenvalues} and also the eigenvectors corresponds to the smaller eigenvalues are also shown in figure \ref{figure:more_eigenvectors} 
only to give you a better idea how the eigenvectors changing in order of their corresponding eigenvalues.

In our experiment, we choose the first five principle components for data projections. This means that we just reduce the dimensionality of the data from 630000($900 \times 600$) dimensions (number of pixels) to only 5
without losing precision which is an enormous amount of compression, processing power saving and memory saving.

Having chosen the number of principle components to keep $\nu$ based on the largest eigenvalues $\lambda$, 
we start writing back the learned PCA to the memory, more specifically we write $\mu, \lambda, \nu$. 
Writing the $\mu, \nu, \lambda = \Lambda^{2}$ to the memory is for the use for recognition module.

\subsection{Recognition of new input data}
Having learned the input camera images, as shown in figure \ref{figure:recognition}, we first read the recorded data $IO$ from the memory using 
the aforementioned recorded XML files in figure \ref{figure:XML_data} using boost library functionality \cite{boost} along with the learned model $\mu, \Lambda, \nu$ from the memory all together 
and then start comparing the new input data $I_{t+1}$ with the input data $I$ one by one from the memory based on the explained four metrics in the recognition part ~\ref{subsec:recognition} 
such as \ref{subsubsec:MNCS}, \ref{subsubsec:MSD}, \ref{subsubsec:SMCS}, \ref{subsubsec:SMSD}.

\begin{figure}
 \centering
 \includegraphics[scale=0.08]{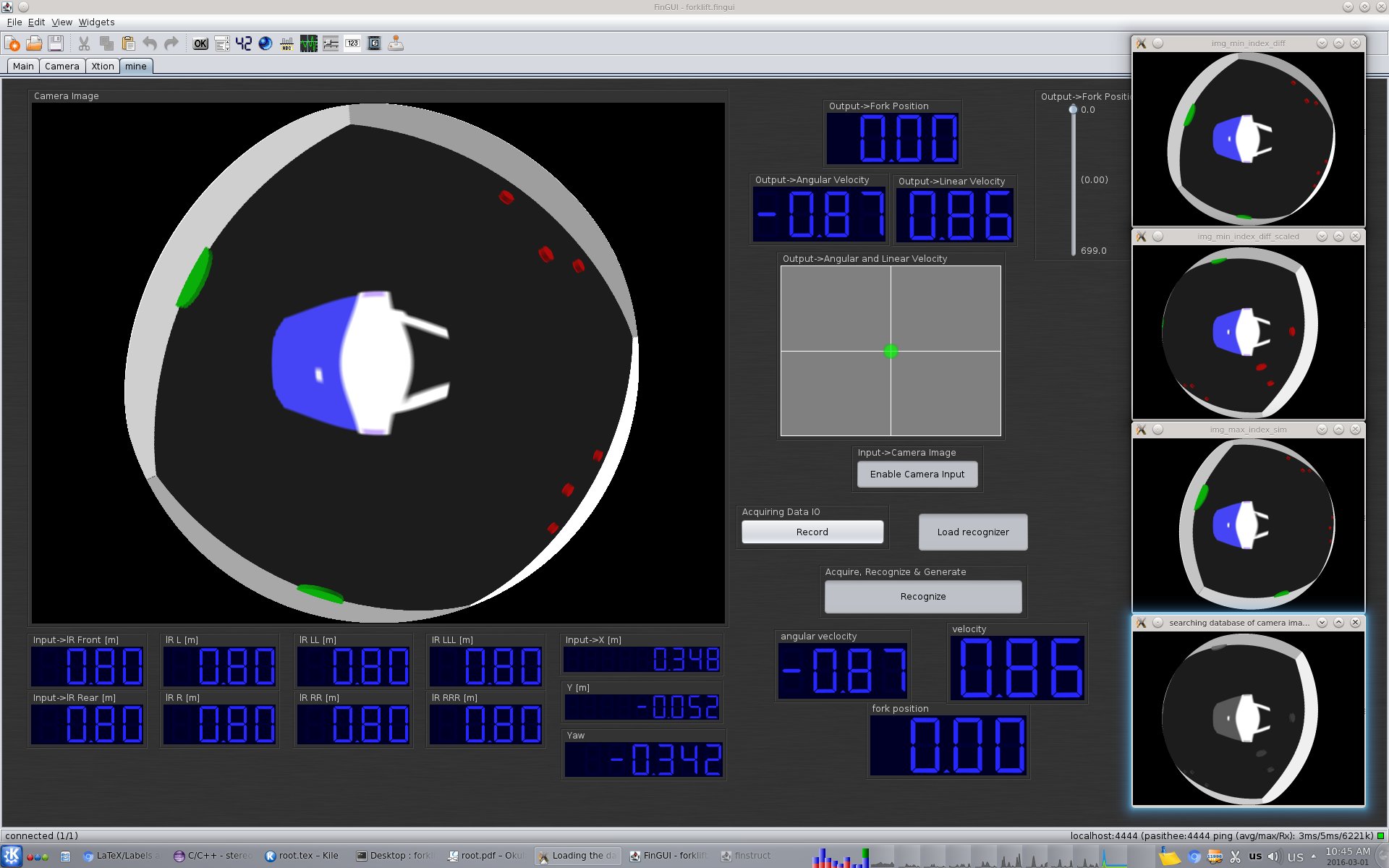}
 \includegraphics[scale=0.08]{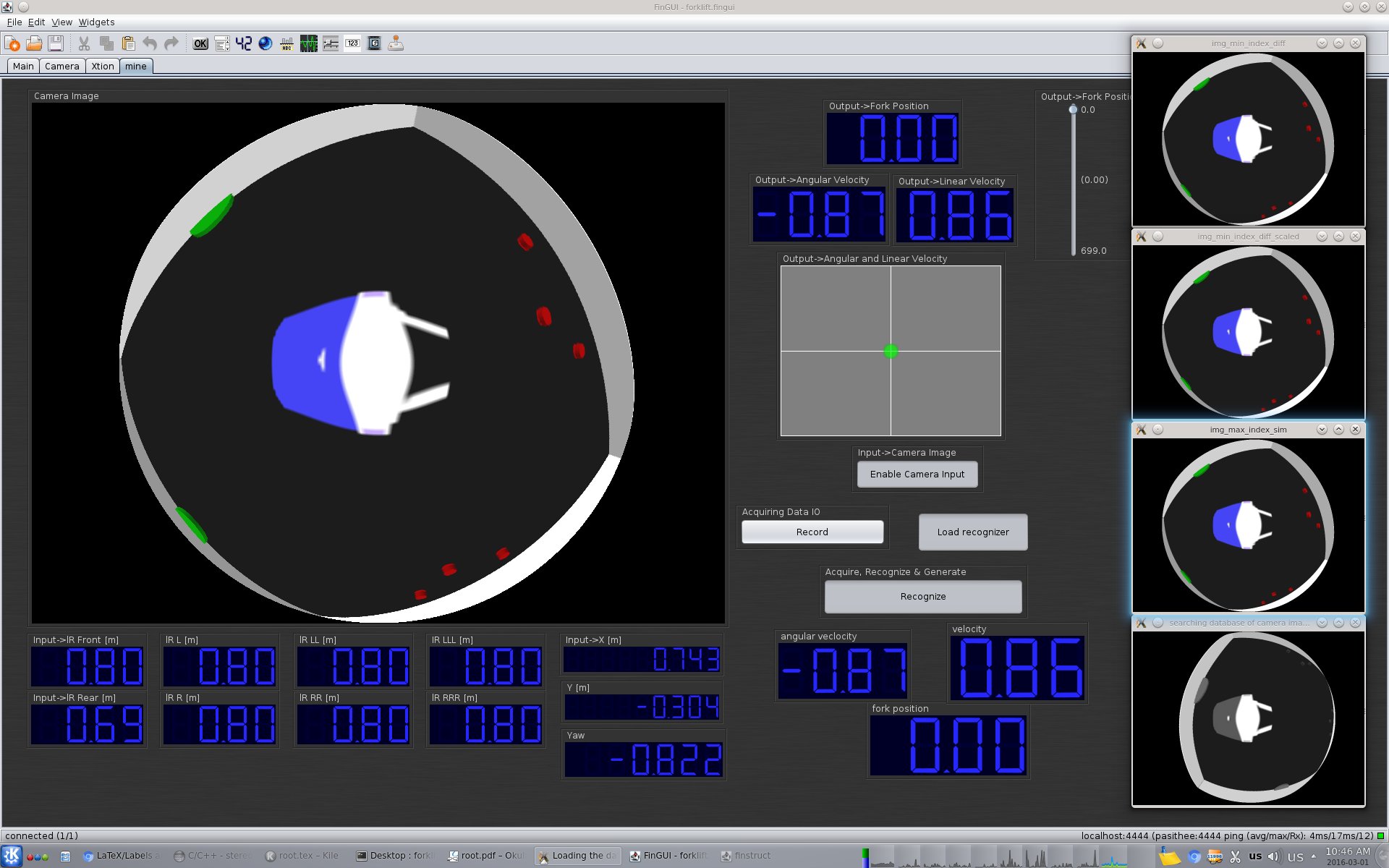}
 \caption{This is how recognizer finds the best match for the input and generates its corresponding output afterwards.}
 \label{figure:recognizer_search}
\end{figure}

\begin{figure}
 \centering
  \includegraphics[scale=0.15]{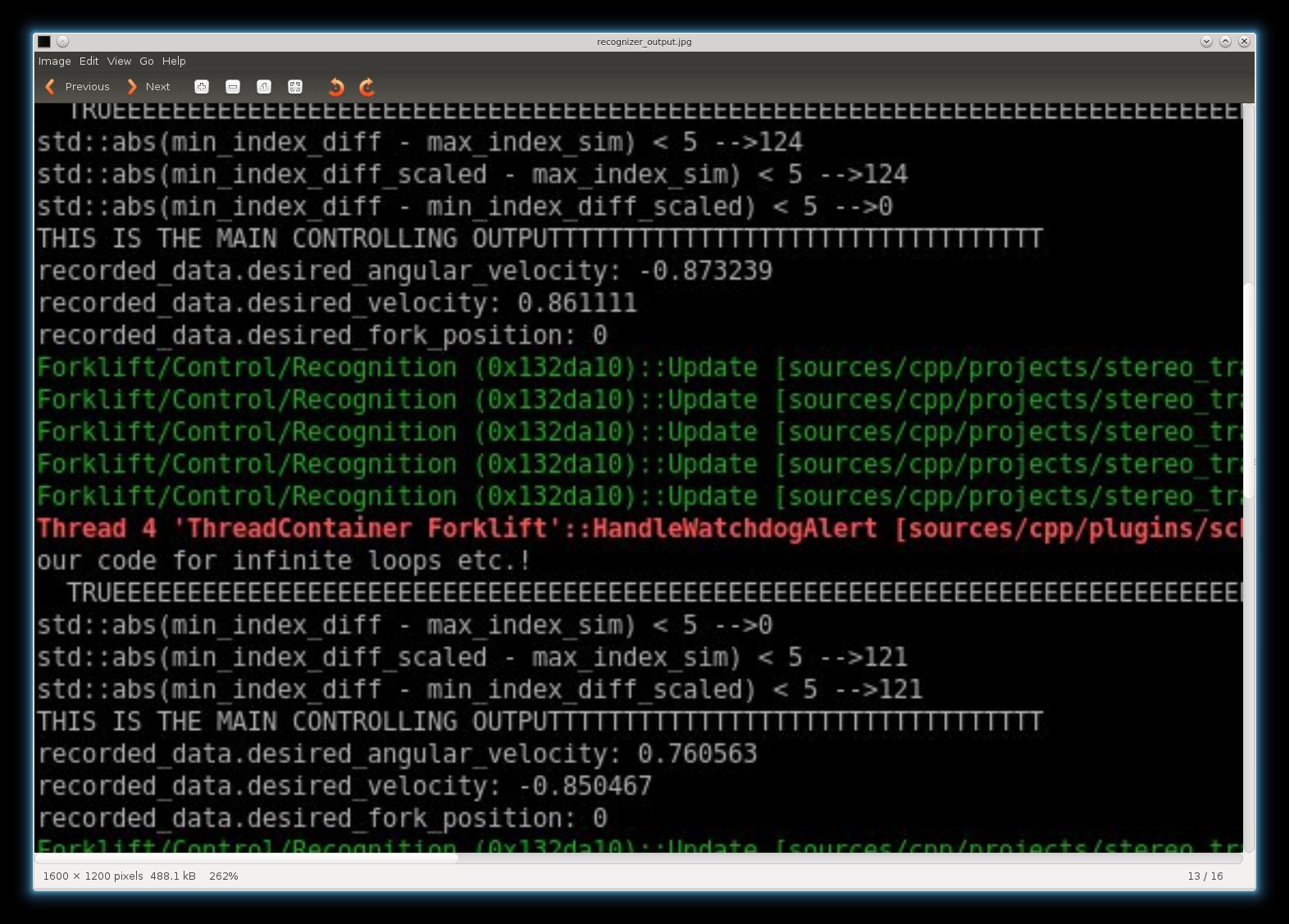}
 \caption{Having found the best match, recognizer generates the corresponding output of the best input in database to the controller input in order to control the robot's action and next move.}
 \label{figure:recognizer_output_zoomed_in}
\end{figure}

In order to start the intelligent robot control, we start the simulation at first, enable the camera images flow then, and load the recognizer with the learned model and the file path of all the recorded XML files 
(\textbf{only the file path}).
Then the robot does an online recognition as shown in figure \ref{figure:recognizer_search} and finally the output is generated for controlling the robot which is the corresponding
linear velocity and angular velocity of the most similar and least different input $I$ in the memory compared to the new input $I_{t+1}$ as shown in figure \ref{figure:recognizer_output_zoomed_in}.

\section{Conclusions and Future Work}

The goal of this paper was introducing a new conceptual cognitive model which may redefine intelligence and
add parameters to intelligence as well as redefining control as a cognition process.
The question addressed is how we can generate a specific detailed intelligence with a strong mathematical foundation and also general enough to be applicable
to any input data for generating any kinds of output data for control.
This is the application of this cognitive model (MLR) in robotics for intelligent control but later on we may also be able to apply this to any other sensor data for any other application as well.

The discussed MLR model is a cognitive process which should be applicable to any kinds of the robot platforms or data structure and data types. 
For this reason, there are plenty of ideas for applying this model to different applications and different sensor data. 
More specifically, in the presented experimental results, we still have distance and localization input data as shown in figure ~\ref{figure:XML_data} which also can be a very interesting quest to apply MLR on them and also their combination as well 
meaning that applying MLR on all of the sensor data fused together.

There is plenty of room and flexibility in putting this concept into an experiment but we should also notice that learning can be very time-consuming, recognition however should be almost real-time. We need to keep that in mind as well.
Also the main power the current model is the ability to handle a high-dimensional data in a way that it reduces the dimensionality without reducing the accuracy of their reconstruction which is a very powerful tool in
handling and managing the big-sized data set of different kinds. That is why one of the important future work will be to apply MLR on a huge database of sensor data and use that for intelligent control.


%

\section*{Acknowledgements}
This research had been conducted in the Robotics Research Lab in the department of Computer Sciences at Univ. Kaiserslautern.




\end{document}